\documentclass{article} 
\usepackage{iclr2025_conference,times}

\usepackage{amsmath,amsfonts,bm}

\def\eqref#1{equation~\ref{#1}}

\def\1{\bm{1}}

\DeclareMathAlphabet{\mathsfit}{\encodingdefault}{\sfdefault}{m}{sl}
\SetMathAlphabet{\mathsfit}{bold}{\encodingdefault}{\sfdefault}{bx}{n}

\usepackage{hyperref}
\usepackage{url}
\usepackage{hyperref}
\usepackage{xcolor}

\hypersetup{
    colorlinks=true, 
    linkcolor=blue,  
    urlcolor=blue,    
    citecolor=black
}
\usepackage{CJKutf8}
\usepackage{xcolor} 

\usepackage{enumerate}
\usepackage{amsthm}
\usepackage{graphicx}
\usepackage{epstopdf}
\usepackage{caption}
\usepackage{color}
\usepackage{enumitem}
\usepackage{array}
\usepackage{tabularx}
\usepackage{multirow}
\usepackage{bm}
\usepackage{booktabs}
\usepackage{subfig}
\usepackage{authblk}

\usepackage[skins]{tcolorbox}
\tcbuselibrary{breakable}
\usepackage{wrapfig}
\usepackage{caption}
\usepackage{subcaption}
\usepackage{listings}
\title{OpenCSG Chinese Corpus: A Series of High-quality Chinese Datasets for LLM Training}

\author[a,b]{Yijiong Yu}

\author[b]{Ziyun Dai}
\author[b]{Zekun Wang}
\author[b]{Wei Wang}
\author[b]{Ran Chen}
\author[b]{Ji Pei}

\affil[a]{Tsinghua University}
\affil[b]{OpenCSG}

\begin{document}

\maketitle

\begin{abstract}

Large language models (LLMs) have demonstrated remarkable capabilities, but their success heavily relies on the quality of pretraining corpora. For Chinese LLMs, the scarcity of high-quality Chinese datasets presents a significant challenge, often limiting their performance. To address this issue, we propose the \textbf{OpenCSG Chinese Corpus}, a series of high-quality datasets specifically designed for LLM pretraining, post-training, and fine-tuning. This corpus includes \textbf{Fineweb-edu-chinese}, \textbf{Fineweb-edu-chinese-v2}, \textbf{Cosmopedia-chinese}, and \textbf{Smoltalk-chinese}, each with distinct characteristics: Fineweb-edu datasets focus on filtered, high-quality content derived from diverse Chinese web sources; Cosmopedia-chinese provides synthetic, textbook-style data for knowledge-intensive training; and Smoltalk-chinese emphasizes stylistic and diverse chat-format data. The OpenCSG Chinese Corpus is characterized by its high-quality text, diverse coverage across domains, and scalable, reproducible data curation processes. Additionally, we conducted extensive experimental analyses, including evaluations on smaller parameter models, which demonstrated significant performance improvements in tasks such as C-Eval, showcasing the effectiveness of the corpus for training Chinese LLMs. The datasets are publicly available on \href{https://huggingface.co/collections/opencsg/high-quality-chinese-training-datasets-66cfed105f502ece8f29643e}{huggingface}.


\end{abstract}

\section{Introduction}
LLM training typically requires vast amounts of data, yet obtaining sufficiently large and diverse corpora for Chinese poses multiple hurdles. For example, \textit{annotation} can be prohibitively expensive and time-consuming, especially when domain experts are required to label specialized content \citep{wang2022supernatkinst}. Robust \textit{infrastructure} is also needed to store and process massive corpora, which demands substantial computational resources \citep{brown2020language}. In addition, \textit{licensing and copyright} concerns complicate the open distribution of textual data \citep{scao2022language}, particularly for non-English material such as Chinese. While prominent Chinese LLMs---including Qwen2 \citep{yang_qwen2_2024}, GLM \citep{du_glm_2022}, and Deepseek \citep{deepseek-ai_deepseek-v2_2024}---have demonstrated strong performance, most rely on proprietary data, limiting community access and slowing advances in Chinese NLP.

Drawing inspiration from \textbf{SmolLM} and \textbf{SmolLM2} \citep{allal2024SmolLM2}, which utilize partially automated pipelines and crowd-sourced content to reduce collection costs, we seek to adopt similarly efficient strategies for assembling large-scale Chinese datasets. At the same time, the \textbf{FineWeb} framework \citep{penedo2024Fineweb-2} offers proven filtering methods that transform raw corpora into clean, context-rich training data. By merging these ideas---cost-effectiveness from SmolLM/SmolLM2 and high-quality filtering from FineWeb---our approach balances practical scalability with rigorous curation to produce openly accessible resources for Chinese LLMs.

Based on these insights, we introduce four datasets, each addressing different aspects of data quality and utility:

\begin{itemize}
    \item \textbf{Fineweb-edu-chinese} and \textbf{Fineweb-edu-chinese-v2} provide large corpora. Both sets rely on a Qwen2-based \citep{yang_qwen2_2024} scorer to emphasize ``educational value,'' ensuring valuable and coherent text. The v1 variant contains about 90 million samples with about 200 billion tokens. The v2 variant doubles the dataset size to over 180 million samples with 420 billion tokens, employing tighter filtering to improve clarity and completeness.

    \item \textbf{Cosmopedia-chinese} is a fully synthetic dataset that mainly comprises textbook-like chapters generated by glm4-longwriter. 
    

    \item \textbf{Smoltalk-chinese} captures multi-turn conversations created through system prompts and advanced Chinese LLMs (e.g., Deepseek-V2.5, Qwen2.5-72b-Instruct). This pipeline forgoes manual data gathering of dialogues but incorporates automated scoring, classification, and de-duplication to mitigate potential style biases or repetitive content.
\end{itemize}

Our method integrates automated scoring modules, synthetic text generation, and domain-focused curation, these \textbf{Chinese datasets} stand apart for their scalability, diversity, and openness. By refining techniques introduced by SmolLM, SmolLM2, and FineWeb, we strive to address recurring obstacles in data collection---such as cost, infrastructure demands, and licensing restrictions---while still ensuring the rich quality and broad coverage essential for training robust Chinese LLMs.

\section{related works}

Pretraining large language models (LLMs) requires vast amounts of text data, yet raw datasets such as CommonCrawl \citep{noauthor_common_nodate} are often noisy and unstructured, making direct training inefficient and less effective. To address this, refined corpora such as C4 \citep{dodge2021documenting}, RedPajama \citep{together2023redpajama}, SlimPajama \citep{cerebras2023slimpajama}, and DCLM-baseline \citep{li2024datacomplm} have been developed. These datasets leverage strict filtering techniques, including heuristic rule-based filtering, trained scorer models, and deduplication methods like MinHash, transforming raw corpora into clean, structured datasets suitable for pretraining. As data processing techniques have advanced, datasets like RefinedWeb \citep{refinedweb}, FineWeb \citep{penedo2024the}, and FineWeb-2 \citep{penedo2024Fineweb-2} have set new standards for data quality, improving both the efficiency and effectiveness of pretraining.

While these advances have significantly improved English datasets, high-quality Chinese corpora remain limited in both scale and refinement. Existing Chinese datasets, such as Wudao-200GB \citep{baai_wudao_2023}, CCI data \citep{wang_cci30-hq_2024}, and Skypile-150B \citep{wei2023skywork}, are relatively coarse, relying on rudimentary processing methods that do not ensure high-quality filtering. This disparity highlights the need for more robust and sophisticated approaches to Chinese dataset curation.

Synthetic data has emerged as a powerful tool for enriching training corpora. Initially demonstrated in instruction tuning through methods like self-instruct \citep{wang2022self} and evol-instruct \citep{luo2023wizardcoder}, synthetic data generation uses powerful LLMs, such as ChatGPT, to create instructions and responses from seed examples, significantly increasing the size of fine-tuning datasets. Magpie \citep{xu2024magpiealignmentdatasynthesis} further streamlined this process by enabling LLMs to directly generate data from system prompts, eliminating the need for seed data and simplifying dataset creation. Beyond fine-tuning, synthetic data has proven effective in pretraining, as evidenced by the Phi series models \citep{gunasekar_textbooks_2023}, which demonstrated its potential to create comprehensive pretraining datasets, though their data remains proprietary. To address this limitation, Cosmopedia \citep{benallal2024Cosmopedia}, the largest open-source synthetic pretraining dataset to date, offers 25 billion tokens of synthetic textbooks, blog posts, stories, and instructional articles generated by Mixtral-8x7B-Instruct-v0.1 \citep{jiang_mistral_2023}. This highlights the scalability and quality achievable with advanced synthetic data generation techniques.

While synthetic data holds great promise, its application to Chinese datasets is still underexplored, largely due to the computational resources required for generation and the challenge of ensuring the data's utility. Existing Chinese fine-tuning datasets also face limitations. For example, llama2-chinese \citep{flagalpha_flagalphallama2-chinese_2023} provides only 10k samples, and Firefly-Train-1.1M \citep{jianxin_yangjianxin1firefly_2024} covers 23 tasks but is restricted to single-turn conversations with short text lengths. ShareGPT-Chinese-English-90k \citep{ShareGPT-Chinese-English-90k} extends to multi-turn conversations but lacks sufficient diversity and quality control. Infinity-Instruction \citep{InfinityInstruct2024} iteratively builds datasets using instruction selection and evolution, yet only 10\% of its millions of samples are in Chinese, limiting its suitability for bilingual or exclusively Chinese training.

The progression from raw corpora to refined datasets like FineWeb underscores the importance of quality control in pretraining data, while advances in synthetic data generation, such as Cosmopedia and Magpie, demonstrate the potential for efficient dataset expansion. However, Chinese datasets still lag behind their English counterparts in scale, quality, and sophistication. Bridging this gap requires combining robust filtering techniques with advanced synthetic data generation methods tailored to the specific challenges of Chinese text. Our work builds on these advancements, integrating the strengths of existing approaches while addressing their limitations, to provide scalable, diverse, and high-quality resources for Chinese LLM research.

\section{Dataset Construction Process}

\begin{figure}[htb]
    \centering
    \hspace*{-0.3cm} 
    \includegraphics[width=1.1\linewidth]{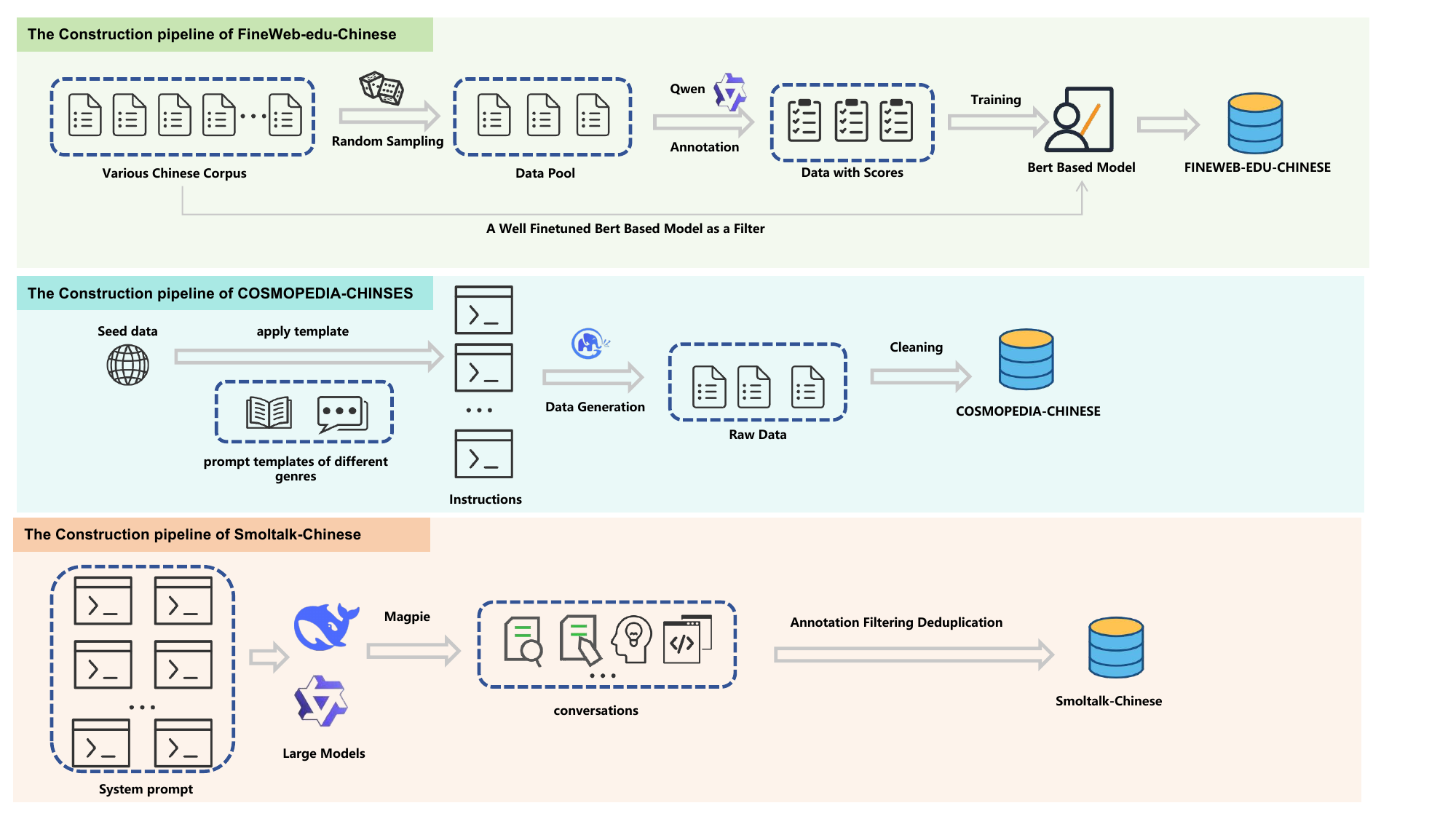}
    \caption{The diagram illustrates the construction pipelines for three Chinese datasets: FineWeb-edu-Chinese, COSMOPEDIA-Chinese, and Smoltalk-Chinese. The FineWeb-edu-Chinese pipeline begins with various Chinese corpora, followed by random sampling, data pooling, annotation, and scoring. A fine-tuned BERT-based model is trained to filter and generate the final dataset. The COSMOPEDIA-Chinese pipeline starts with seed data collection, proceeds through prompt design and data generation, and results in a database of curated knowledge. Lastly, the Smoltalk-Chinese pipeline leverages powerful Chinese LLMs with task-specific system prompts to generate conversational data.}
    \label{fig:fineweb}
\end{figure}

This chapter introduces the construction process of three datasets designed to meet the demand for high-quality and domain-relevant Chinese data. Inspired by previous works such as Fineweb-edu \citep{penedo2024the} and Cosmopedia-v2 \citep{benallal2024Cosmopedia}, we tailored and expanded upon their methodologies to address the challenges of limited Chinese data availability and to ensure the datasets meet diverse task requirements.

\subsection{Fineweb-Edu-Chinese Dataset}

The Fineweb-Edu-Chinese dataset construction pipeline largely follows the methodology of Fineweb-edu. However, unlike Fineweb-edu, which is filtered from the 15-terabyte Fineweb corpus, large-scale Chinese data is relatively scarce. To address this, we assembled multiple open-source Chinese corpora, including Wudao \citep{baai_wudao_2023}, Telechat \citep{wang2024telechat}, Map-CC \citep{du2024chinesetinyllmpretraining}, CCI2 \citep{wang_cci30-hq_2024}, and Skypile \citep{wei2023skywork}, to form the original data pool. These datasets were selected for their diversity and their educational and technical relevance.

From the CCI2 dataset, we randomly sampled one million entries and used Qwen2-7b-instruct \citep{yang_qwen2_2024} to evaluate the educational value of each sample on a scale from 0 to 5, using the prompt shown in Appendix \ref{app:fineweb}. This scoring data was then used to fine-tune a bge-rerank-zh model \citep{zhang_bge_2023}, enhanced with a linear regression layer. The model was trained for three epochs using a cosine decay learning rate of 1e-4, balancing computational efficiency and model performance. Using this scorer, the dataset was filtered to exclude samples with scores below 3, retaining high-quality entries.

To remove redundancy, we applied Min-Hash with an overlap threshold of 0.7. This technique ensured computational efficiency while maintaining diversity in the dataset. The resulting Fineweb-Edu-Chinese dataset consists of 89 million high-quality samples, offering a rich resource for educational and technical applications.

\subsection{Fineweb-Edu-Chinese-V2 Dataset}
In the V2 version of Fineweb-edu-chinese, we expand the data pool, adding Michao \citep{liu2023michaohuafen10specializedpretrained}, CCI3 \citep{wang_cci30-hq_2024}, IndustryCorpus2 \citep{industryCorpus2} and ChineseWebText \citep{chen2023chinesewebtext}. And in the first step, we replace Qwen2-7b-instruct with Qwen2.5-14b-instruct \citep{team_qwen25_2024}, and one million samples is sample from the entire pool. The other steps are almost the same.

\subsection{Cosmopedia-chinese}

Inspired by Cosmopedia-v2 \citep{benallal2024Cosmopedia}, we extended the dataset construction process by generating synthetic textbook-like data from seed samples. Seed data plays a critical role in guiding the generation of high-quality samples, which is expected to contain valuable knowledge. While web texts from the Chinese data pool were initially considered, their quality was not high enough, as they often consisted of advertisements or lacked valuable knowledge.

To overcome this, we collected seed samples from high-quality sources, including 5.6 million entries from BaiduBaike, 1 million samples from Zhihu Q\&A, and 2 million entries from technical blogs. These sources were selected for their rich domain knowledge and high informational density. 

Since the experiments of Cosmopedia-v2 have found that using larger models, such as 70B level LLMs, to generate data did not provide no significant improvements, we decided to just choose 7B level models to accelerate the synthetic process. However, after trying qwen2-7b-instruct \cite{yang_qwen2_2024} or yi-1.5-9b-chat \cite{ai2024yiopenfoundationmodels}, we find usual chat models always tend to output concise and generalizable content like an abstract or outline, rather than sufficiently detailed and specific content like the main content of the textbook, even though we have prompted the model to answer as detailed as possible. Thus, we resort to glm4-9b-longwriter \cite{bai2024longwriter}, which can generate long answers with sufficiently detailed content.

Using glm4-9b-longwriter \citep{bai2024longwriter}, we generated synthetic samples in various genres, such as textbook units, narrative stories, and detailed ``how-to" guides. The prompts we used for generating each genre are in Appendix \ref{app:cosmo}. A temperature setting of 0.8 ensured diverse outputs. From 20 million generated samples, deduplication via Min-Hash retained 15 million high-quality entries, preserving diversity without compromising data quality.

\subsection{Smoltalk-Chinese Dataset}

The Smoltalk-Chinese dataset builds on the methodologies of Magpie-ultra-1M \citep{xu2024magpiealignmentdatasynthesis} and Smoltalk \citep{allal2024SmolLM2}, addressing gaps in task diversity and conversational depth. Since Smoltalk includes some extra tasks besides Magpie-ultra-1M which focusing on 11 task categories, we also introduces 7 additional task categories in Smoltalk-Chinese: format constraints, summarization, rewriting, document-QA, safe-QA, translation, and everyday-talk. This expansion ensures broader coverage of tasks relevant to natural language understanding and generation.

Using advanced LLMs such as Deepseek-V2.5 \citep{deepseek-ai_deepseek-v2_2024} and Qwen2.5-72B-Instruct \citep{team_qwen25_2024}, we generate 3-turn conversations for the 11 task categories used in Magpie-ultra-1M, 1-turn conversations for the new categories except everyday-talk, and 5-turn conversations for everyday-talk. The system prompt and detailed description of each task category is shown in Appendix \ref{app:magpie}. To ensure diversity, which is highly important for instruction-tuning, we set the temperature to 1.2.

To enhance quality, Qwen2.5-7b-instruct was used to score the first user query of each generated conversation, based on its clarity and coherence, and we only keep those with the score over 3. The prompts used to instruct Qwen2.5-7b-instruct is shown in Appendix \ref{app:magpie}.

Deduplication was performed using embeddings encoded by gte-zh-large \citep{zhang2024mgte} and filtered based on cosine similarity thresholds of 0.8 for multi-turn samples and 0.7 for single-turn samples. After filtering, the Smoltalk-Chinese dataset contains about 70,000 high-quality samples, offering a robust resource for tasks requiring diverse conversational and task-oriented data.

\section{Experiments and Analysis}

This chapter presents the experiments conducted to validate the effectiveness of our three proposed datasets—\textbf{Fineweb-Edu-Chinese}, \textbf{Cosmopedia-Chinese}, and \textbf{Smoltalk-Chinese}—in pretraining and fine-tuning language models (LMs). We focus on systematically assessing each dataset’s contribution to model performance, comparing to other open-source datasets, and drawing insights into potential areas of improvement. 


\subsection{Fineweb-Edu-Chinese}


\begin{figure}
    \centering
    \includegraphics[width=0.4\linewidth]{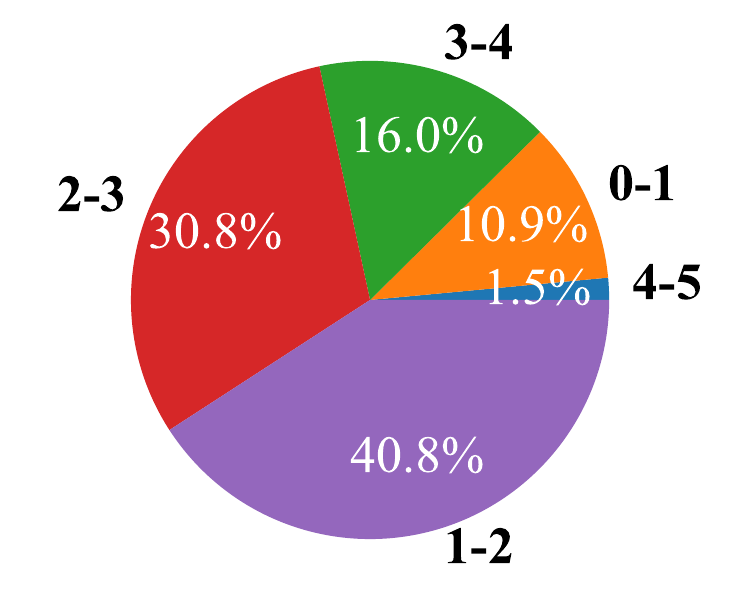}
    \caption{Score distribution of all the unfiltered source data scored by the Fineweb-Edu-Chinese-v2 scorer. High-quality samples (score $>3$) form only a small fraction, indicating the scarcity of valuable data in open-source Chinese corpora.}
    \label{fig:score dist}
\end{figure}

\begin{figure}[htb]
    \centering
    \subfloat[Text length]{\includegraphics[width=0.5\linewidth]{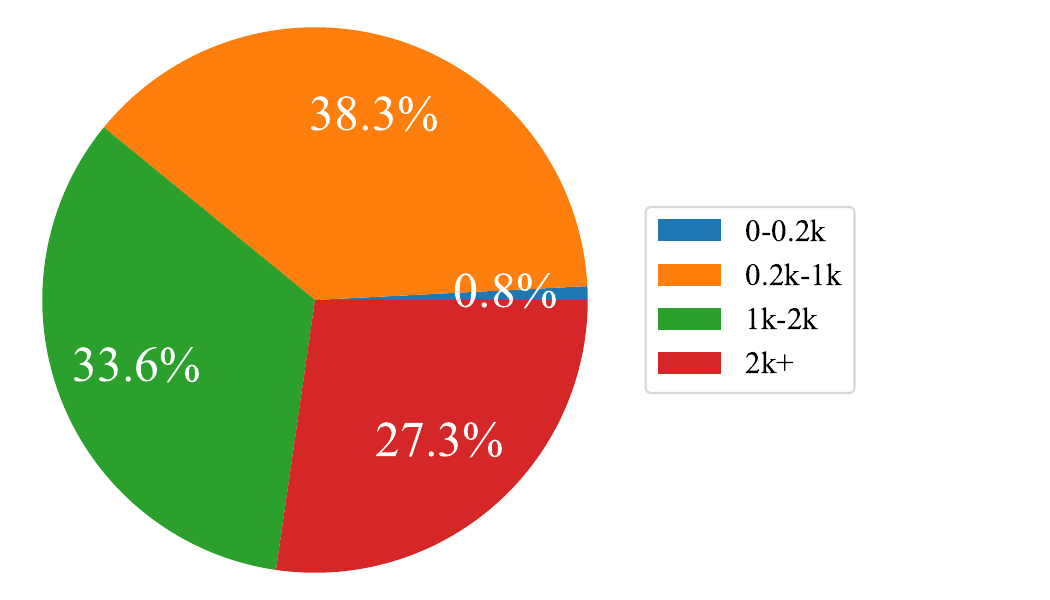}}
    \subfloat[Source]{\includegraphics[width=0.5\linewidth]{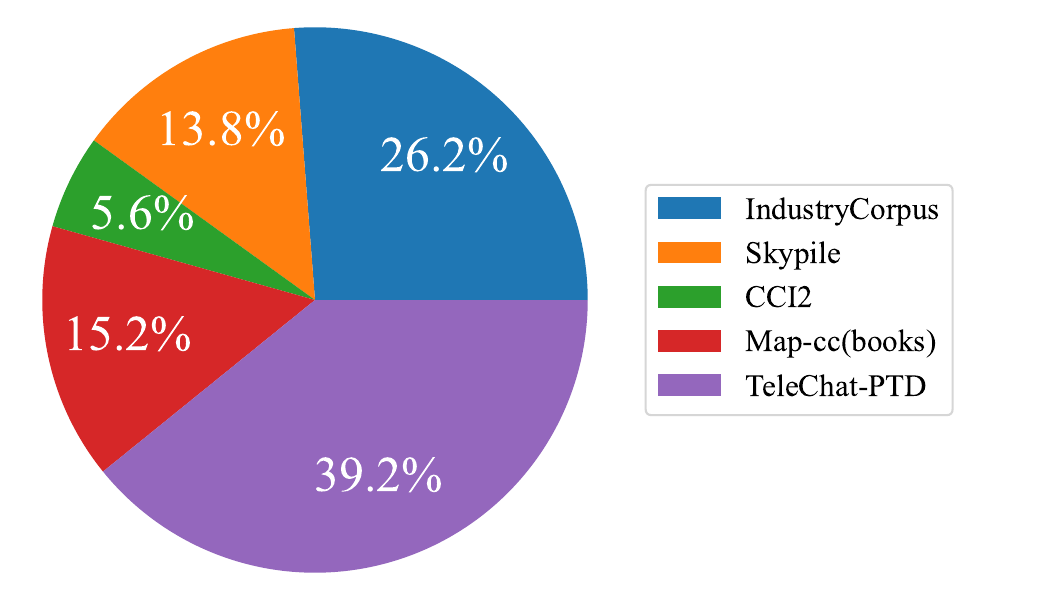}}

    \caption{(a) The text length distribution of Fineweb-Edu-Chinese shows most samples' lengths are in the interval 0.2k-1k and 1k-2k. (b) The source where the samples are from in Fineweb-Edu-Chinese.}
    \label{fig:fineweb v1 stat}
\end{figure}

\begin{figure}[htb]
    \centering
    \subfloat[Text length]{\includegraphics[width=0.5\linewidth]{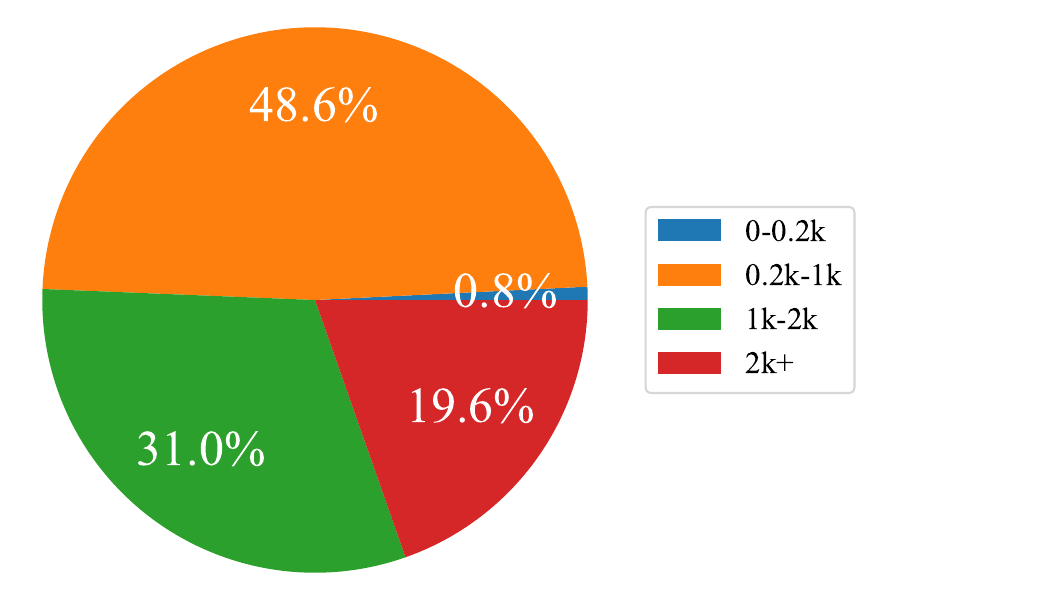}}
    \subfloat[Source]{\includegraphics[width=0.5\linewidth]{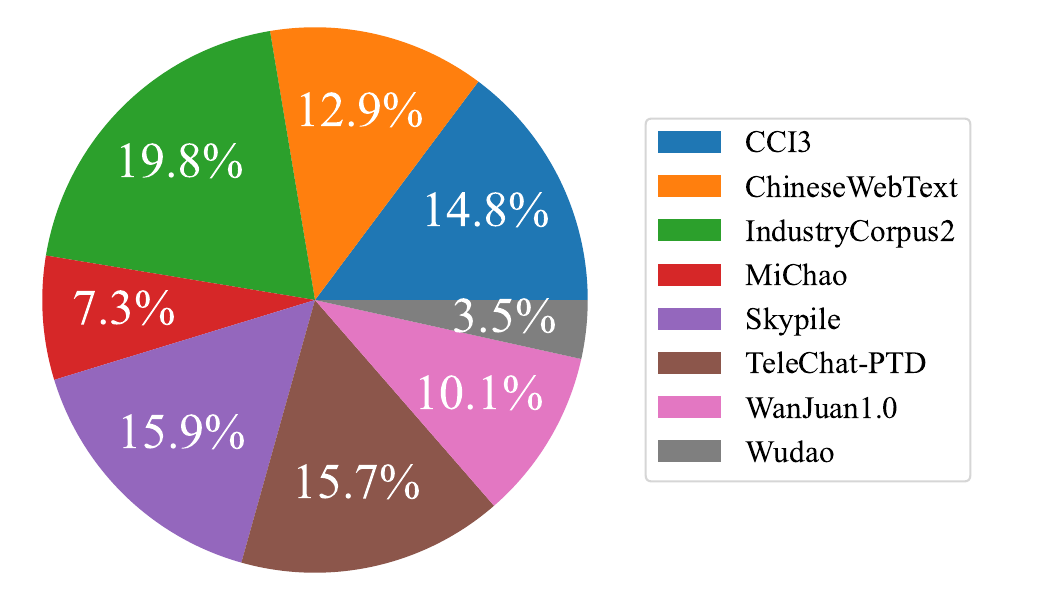}}

    \caption{(a) The text length distribution of Fineweb-Edu-Chinese-v2 shows most samples' lengths are iin the interval 0.2k-1k. (b) The source where the samples are from in Fineweb-Edu-Chinese-v2.}
    \label{fig:fineweb v2 stat}
\end{figure}

We show the score distribution of the the unfiltered data scored by the Fineweb-Edu-Chinese-v2 scorer in Figure~\ref{fig:score dist}. It is clear that most samples from the original Chinese corpora were scored in the 1-2 range, confirming that truly high-quality data is comparatively rare. Despite this low proportion of top-tier samples, careful filtering and rigorous scoring resulted in the Fineweb-Edu-Chinese dataset, which is expected to contain more meaningful and instructional text than typical open-source alternatives.

The statistical information on text length and data sources of Fineweb-Edu-Chinese v1 and v2 is shown in Figure \ref{fig:fineweb v1 stat} and Figure \ref{fig:fineweb v2 stat}.


To prove the dataset's advantage, we pretrained a \textbf{2B}-level language model based on the Llama architecture from scratch using Fineweb-Edu-Chinese v1. A baseline dataset, randomly sampled from the raw source of Fineweb-Edu-Chinese and matched in size, served as our control condition. Both models were trained for over 50k steps using a sequence length of 2048, a global batch size of 512, and a constant learning rate of $1\mathrm{e}{-3}$. We chose a large sequence length (2048) to exploit LLMs’ capacity for capturing extended dependencies, and a relatively high learning rate to accelerate early-stage convergence under limited compute resources.
\begin{figure}[htb]
    \centering
    \subfloat[Evaluation on CMMLU]{\includegraphics[width=0.45\linewidth]{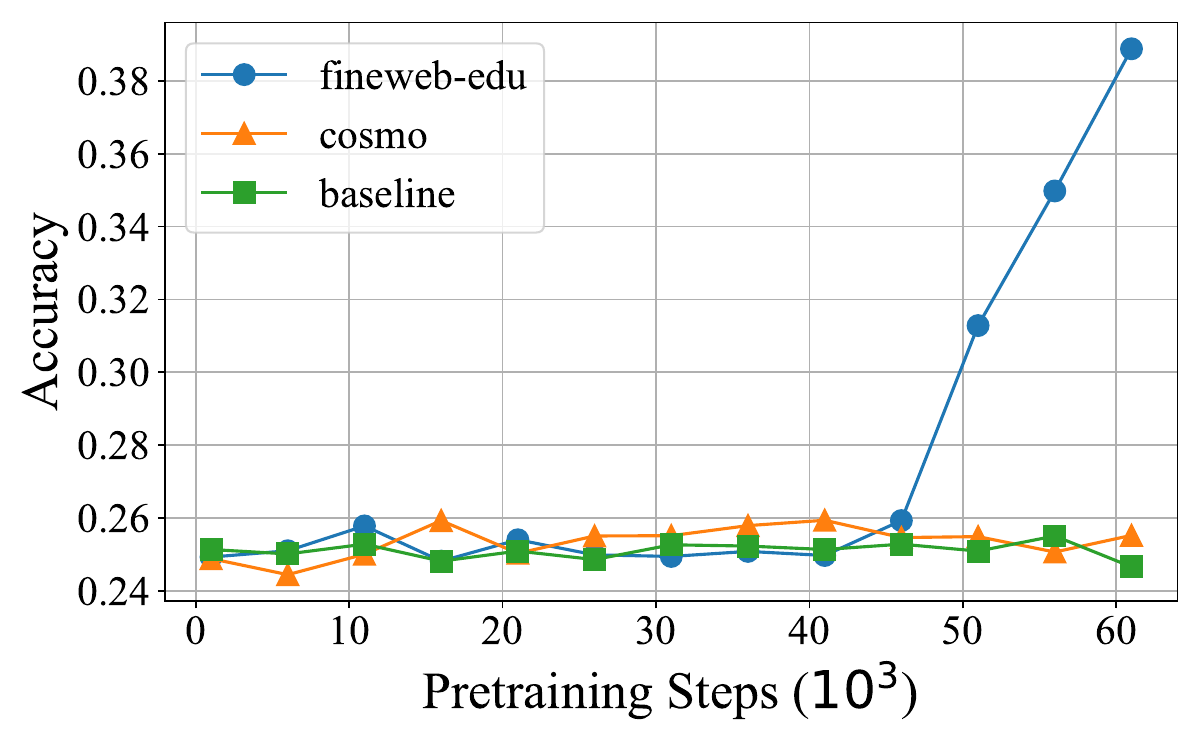}}
    \subfloat[Evaluation on CEval]{\includegraphics[width=0.45\linewidth]{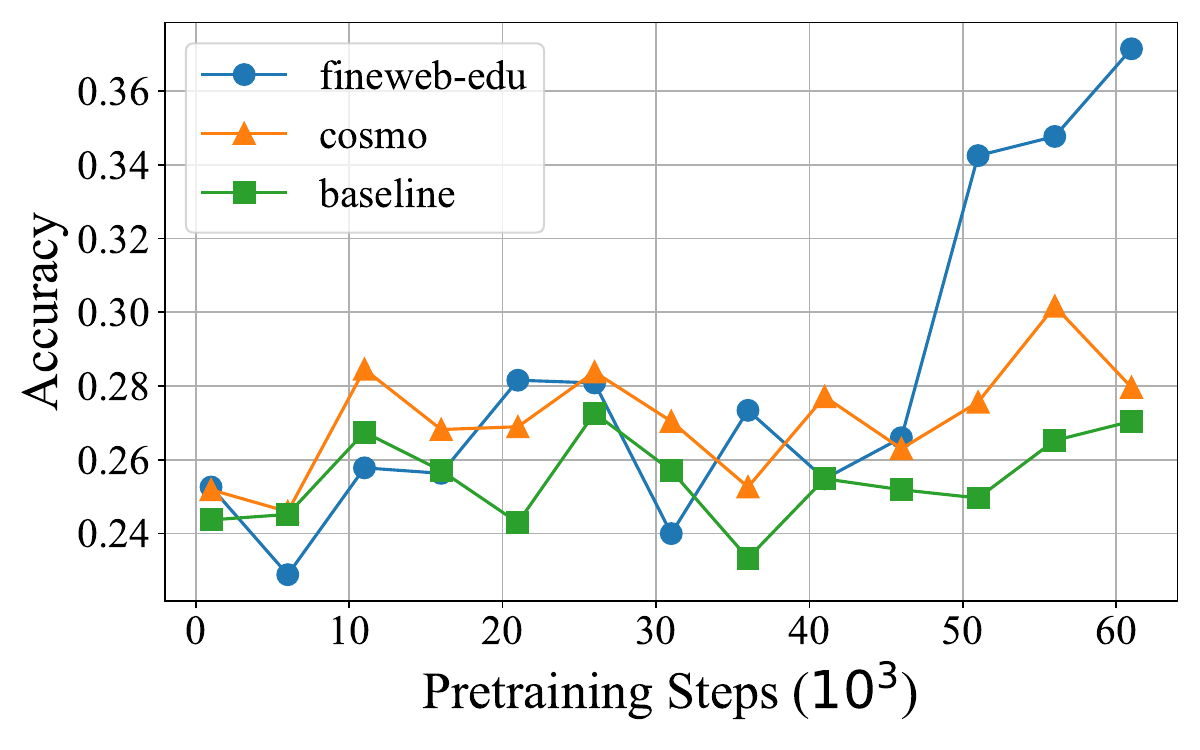}
}
    \caption{The score in CMMLU and CEval of each checkpoint when training with different datasets.}
    \label{fig:3 model acc}
\end{figure}

We assessed model performance on the \textbf{C-Eval}~\citep{huang2023ceval} and \textbf{CMMLU}~\citep{li2023cmmlu} benchmarks, two authoritative suites for evaluating Chinese NLP models’ world-knowledge and understanding abilities. During evaluation, we set the temperature to 0 (for deterministic outputs) and used a 5-shot prompting strategy to provide limited contextual examples. As illustrated in Figure~\ref{fig:3 model acc}, the Fineweb-Edu-Chinese model experienced a sharp accuracy increase around 45k steps, surpassing the baseline by a significant margin. This sudden rise likely reflects the model’s acquisition of domain-specific patterns present in the filtered dataset, confirming that Fineweb-Edu-Chinese’s focus on higher-scoring educational content enhances pretraining efficiency and downstream performance.

Due to the limitation of computational resource, we have not yet run pretraining experiments on Fineweb-Edu-Chinese v2. But due to its larger scale and refined construction pipeline, we are sure it will provide a much better effect. 

\subsection{Cosmopedia-Chinese}

\begin{figure}[htb]
    \centering
    \subfloat[Genre]{\includegraphics[width=0.33\linewidth]{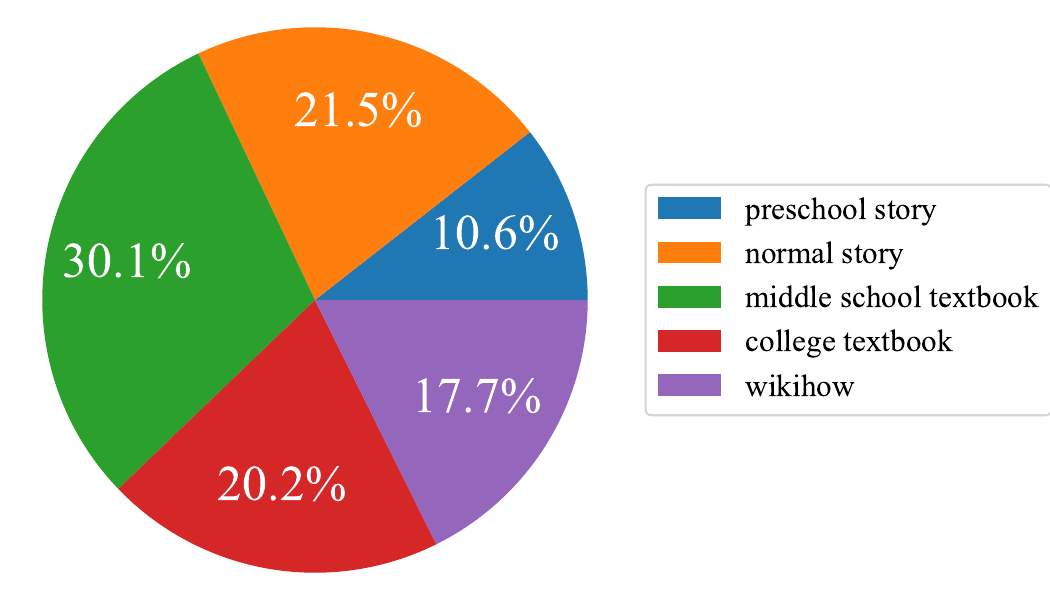}}
    \subfloat[Seed Data]{\includegraphics[width=0.33\linewidth]{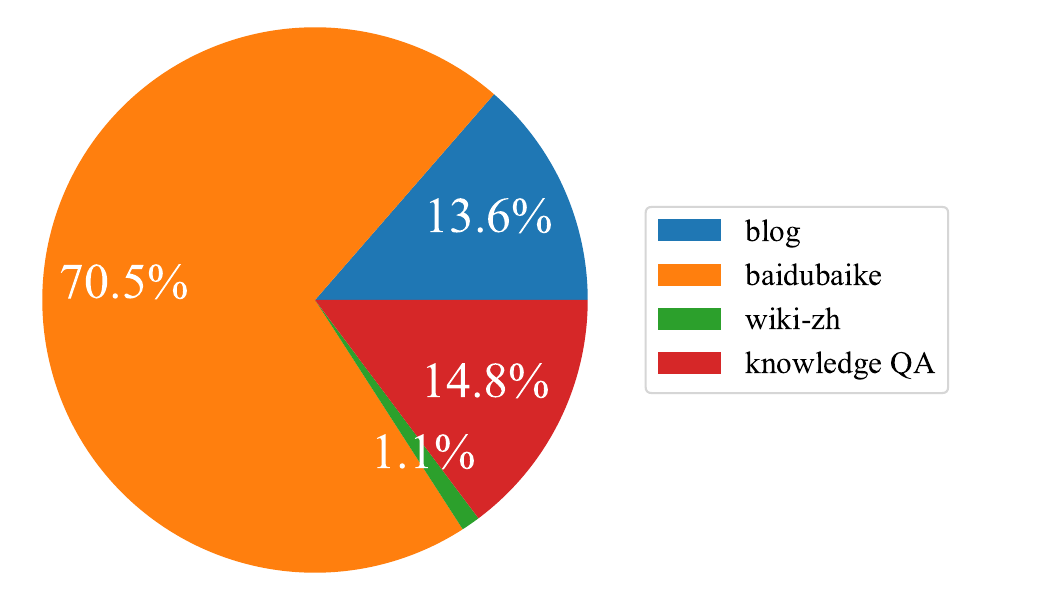}}
    \subfloat[Text Length]{\includegraphics[width=0.33\linewidth]{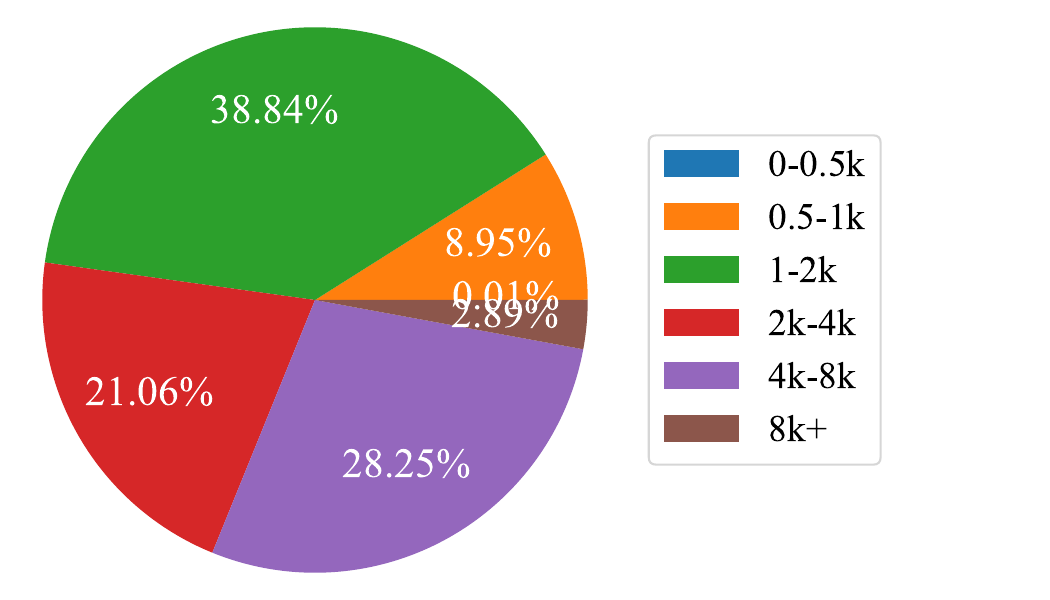}
}

    \caption{Statistic information about Cosmopedia-Chinese.}
    \label{fig:cosmo type dist}
\end{figure}

Cosmopedia-Chinese, in contrast, is synthesized in its entirety from high-quality seed data sources, including BaiduBaike entries, Zhihu Q\&A discussions, and technical blog articles. And each piece of seed data can be used to generate multiple diffirent genres. As illustrated in Figure~\ref{fig:cosmo type dist}, this approach yields a seemingly balanced mix of content genres, suggesting coverage of diverse topics such as science, literature, and practical know-how. And the average text length is relatively longer than normal datasets, with over a half samples are longer than 2k. To rigorously test its effectiveness, we pretrained a \textbf{2B}-level Llama-based model on Cosmopedia-Chinese with the same hyper-parameters used for Fineweb-Edu-Chinese, ensuring a fair comparison.

Despite the dataset’s breadth of topics, benchmark results indicated only a little accuracy gains over the baseline after 2 epochs of training. We attribute this lack of improvement primarily to two factors. First, the \textit{homogeneity of synthesized data}—even though the topics were varied, the limited genres may have led to repetitive rhetorical structures or stylistic patterns that reduced the overall diversity LLMs rely on to achieve robust generalization. Second, the \textit{overuse of markdown formatting} throughout many samples could divert the model’s attention toward parsing formatting tokens rather than absorbing substantive content, potentially diminishing its ability to capture key semantic and contextual signals.

Nonetheless, human evaluators noted that the Cosmopedia-Chinese model produced consistently well-structured and knowledge-rich responses, indicating it effectively inherited some valuable patterns from the seed data. This outcome highlights the dataset’s potential utility in use cases where coherent, polished text is prioritized—such as educational materials, guided tutorials, or domain-specific reference resources. To further improve generalization, we recommend blending \textit{real-world data} with the synthesized corpus and reducing or removing markdown tags during preprocessing, allowing the model to concentrate more fully on semantic richness and contextual variety.
\subsection{Smoltalk-Chinese}

\begin{figure}
    \centering
    \includegraphics[width=0.5\linewidth]{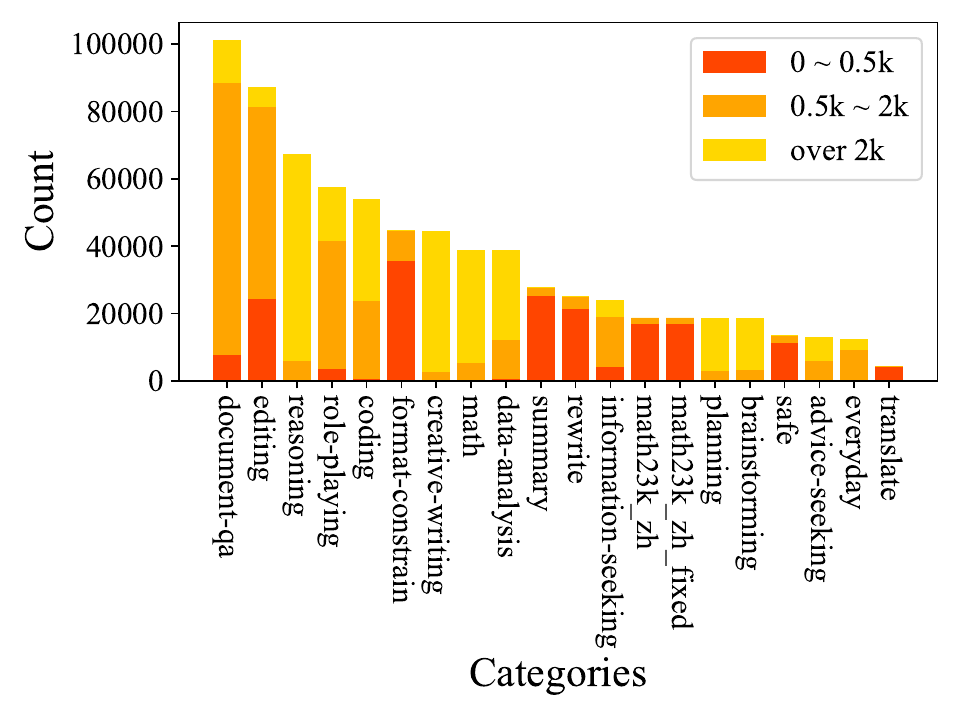}
    \caption{The amount and text-length distribution of each task in Smoltalk-Chinese.}
    \label{fig:smoltalk dist}
\end{figure}

\begin{figure}[htb]
    \centering
    \includegraphics[width=0.7\linewidth]{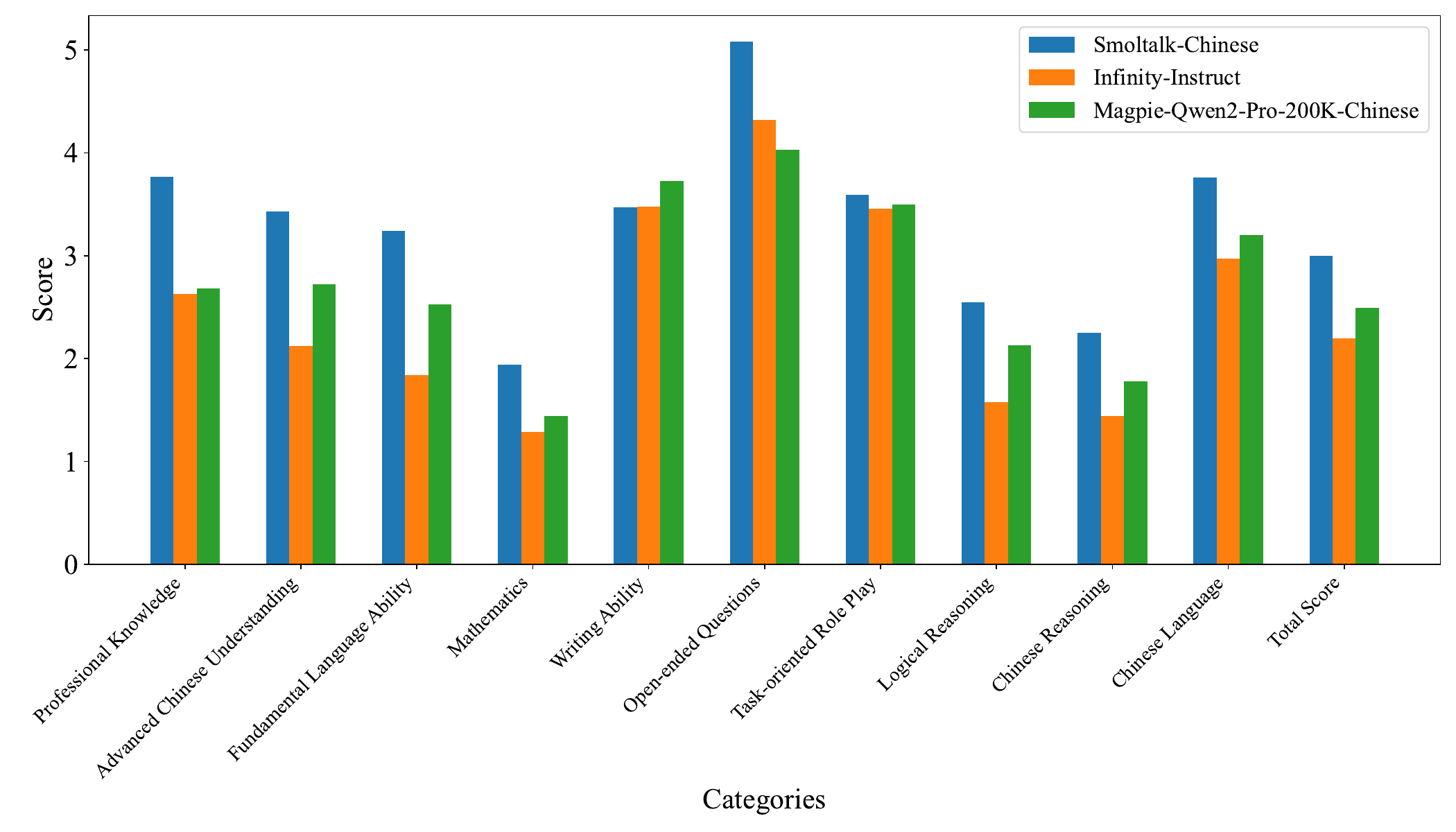}
    \caption{Models' performance on Alignbench after fine-tuned on different datasets.}
    \label{fig:sft score}
\end{figure}

Smoltalk-Chinese is specifically designed for instruction fine-tuning and contains diverse tasks and conversation lengths (Figure~\ref{fig:smoltalk dist}). To assess its effectiveness, we took the 2B model pretrained on Fineweb-Edu-Chinese as our backbone, then fine-tuned it on Smoltalk-Chinese, Infinity-Instruct~\citep{InfinityInstruct2024} (using 100k Chinese-only samples), and Magpie-Qwen2-Pro-200K-Chinese~\citep{xu2024magpiealignmentdatasynthesis}. We conducted training over 2 epochs, with a starting learning rate of $3\mathrm{e}{-4}$ and cosine decay. 

We then evaluated each fine-tuned model on \textbf{Alignbench}~\citep{liu2023alignbench}, a benchmark focused on multi-dimensional alignment criteria for Chinese LLMs, including correctness, helpfulness, clarity, safety, fairness, etc. As shown in Figure~\ref{fig:sft score}, the Smoltalk-Chinese dataset yielded the strongest overall performance gains. Its collection of multi-turn dialogues and broad range of tasks allowed the model to handle sophisticated instructions involving complex reasoning and conversational nuance. 


\section{Conclusion and Limitations}

We propose four Chinese datasets—\textbf{Fineweb-Edu-Chinese-v1}, \textbf{Fineweb-Edu-Chinese-v2}, \textbf{Cosmopedia-Chinese}, and \textbf{Smoltalk-Chinese}—that collectively address pressing needs in Chinese LLM development. Fineweb-Edu-Chinese showed excellent utility for pretraining, significantly boosting downstream performance thanks to its careful filtering of high-value educational content. Cosmopedia-Chinese did not produce strong benchmark gains yet proved capable of generating coherent, knowledge-rich outputs under human assessment. Smoltalk-Chinese excelled in aligning model behaviors with user instructions and exhibited robust improvements across diverse alignment metrics.

OpenCSG Chinese Corpus enhance the Chinese NLP open-source community by broadening both the quality and diversity of available data. Nevertheless, potential improvements remain. The Cosmopedia-Chinese dataset could benefit from real-world data blending to mitigate over-homogeneity and from further reductions in markdown tags. Further research should also explore alternative or complementary evaluation metrics to capture a broader spectrum of model strengths and weaknesses, including factual correctness, reasoning depth, and safety alignment. By addressing these limitations, future iterations of these datasets can better support the development of contextually aware, high-performing large language models in Chinese.

\bibliography{iclr2025_conference}
\bibliographystyle{iclr2025_conference}

\appendix
\section{Prompts}
\subsection{prompts for chinese-fineweb-edu}
\label{app:fineweb}

In the first step of constructing chinese-fineweb-edu, we randomly sampled a million samples from this pool and use Qwen2-7b-instruct to score each sample, based on the following prompt of ranking standard:

\begin{tcolorbox}[colback=white,breakable]\begin{CJK*}{UTF8}{gbsn}{\small
以下是一段网页内容摘录。请使用以下5分制评分系统来评估该网页的写作水平、教育价值和实用性:\\
0分：如果网页没有提供任何教育价值,完全由无关信息(如广告、宣传材料、少儿不宜内容)组成。\\
1分：如果网页提供了一些可能有教育价值的基本信息,即使包含一些无关或非学术内容(如广告和宣传材料)。\\
2分：如果网页涉及某些与教育相关的元素,但与教育标准不太吻合。它可能将教育内容与非教育材料混杂,对潜在的有用的主题进行浅显概述,或以不连贯的写作风格呈现信息。\\
3分：如果网页适合教育使用,并介绍了与某些学校课程中可能学到的关键概念，或对个人发展有用的实用信息。它的内容连贯但可能不全面,或包含一些无关信息。它可能类似于教科书的一小段节选,可以学习但有明显局限,如涉及过于复杂的概念、过于具体的不重要事件。\\
4分：如果网页与教育高度相关，对个人学习发展有益,表现出清晰一致的写作风格。它可能类似于教科书的一个章节或教程,提供大量教育内容,极少包含无关信息,且概念对学生来说不会过于深奥。内容连贯、重点突出,对结构化学习有价值。\\
5分：如果网页摘录在教育价值上表现极好,完全适合小学、中学或大学教学或专业人士学习。它遵循详细的推理过程,写作风格易于理解,对主题提供深刻而全面的见解,不包含任何非教育性或无实用意义内容。\\

网页内容摘录:\\
\{data\}

在审查这段网页摘录后：请简要地为您的评分进行合理的解释，最多不超过100字，最后以“教育得分：【分数】”的格式结束。请根据所列出的标准系统地赋予分数。\\
}
\end{CJK*}
\end{tcolorbox}

\subsection{prompts for chinese-cosmopedia}
\label{app:cosmo}
To construct chinese-cosmopedia, we use the following prompts for generating each style of content:

\begin{tcolorbox}[colback=white,breakable]\begin{CJK*}{UTF8}{gbsn}{\small
\textbf{Textbook for college student}\\
这是一段来自网页的摘录：\\
“\{data\}”\\
请编写一个针对大学生的足够详细的教科书课程单元，该单元与给定的摘录中的某个概念或多个概念相关。\\
不需要包含摘录中的所有内容，只需要发掘其中适合作为教科书内容的部分。你可以自由补充其他相关知识。\\
不能仅仅列出概念，而是要深入发展和详细探讨每个概念，因为我们优先考虑深入理解主题内容，而不是广度。\\
要求：1. 严谨性：确保对概念/章节的深入覆盖。\\
2. 吸引性：用学术、专业且引人入胜的语气撰写，以吸引兴趣。\\
3. 应用：融入具体的实践例子，例如微积分中要给出公式、严格证明，历史中要给出关键日期和人物，计算机操作中要给出代码。\\
4.不需要给出参考文献。内容中不应包含广告或涉及隐私的信息。注重主体内容，不需要其它格式化的内容。\\
请记住，要针对大学生制作内容，他们可能拥有一些基础知识，但不是该领域的专家。内容应该详细且发人深省。\\
请立即开始撰写教科书，不要使用图片，不要输出除了教科书以外的内容，不要以“课程单元”作为标题而是要有具体的标题。

\tcbline
\textbf{Textbook for high school student}\\
网页摘录：
“\{data\}”\\
创建一个与上述网页摘录中的某个概念相关的具有教育意义的内容，针对中学生，尽量长而详细。你可以自由补充其他相关知识。\\
不能仅仅列出概念，而是要深入发展和详细探讨每个概念，因为我们优先考虑深入理解主题内容，而不是广度，不需要包含摘录中的所有内容。\\
不应该使用像微积分这样的复杂大学级主题，因为这些通常不是中学的内容。\\
如果主题是关于这些的，寻找一个更简单的科学替代内容来解释，并使用日常例子。\\
例如，如果主题是“线性代数”，你可能会讨论如何通过将物体排列成行和列来解决谜题。\\
避免使用技术术语和LaTeX，只讨论中学级别的主题。内容中不应包含广告或涉及隐私的信息。\\
请直接开始撰写教育内容，不要输出除了教育内容以外的内容。
\tcbline
\textbf{Wonderful story}\\
写一个与以下文本片段相关的引人入胜的故事：\\
“\{data\}”\\
故事不需要提及片段中的所有内容，只需使用它来获得灵感并发挥创意！可以加入其它知识。故事需要以第一人称叙述，模仿一个人在知乎上分享自己的故事。\\
故事应包括：\\
1.意想不到的情节转折或引人入胜的冲突。 \\
2.反思和洞察：以具有教育意义的新理解、启示的结论结束。 \\
3.请勿包含广告或涉及隐私的信息。\\
请马上开始讲故事，不要输出除了故事以外的内容。

\tcbline
\textbf{Story for baby}\\
网页摘录：
“\{data\}”\\
创建一个与上述网页摘录中的某个概念相关的具有教育意义的儿童故事，重点针对对世界和人际交往零知识的5岁儿童。\\
故事不需要提及片段中的所有内容，只需使用它来获得灵感并发挥创意。\\
故事应该使用简单的术语。你可以补充额外的知识来帮助理解。\\
使用易于理解的示例，并将 5 岁儿童可能提出的问题及其答案纳入故事中。故事应涵盖日常行为和常见物品的使用。\\
不应该使用像微积分这样的复杂大学级主题，因为这些通常不是幼儿能理解的内容。如果主题是关于这些的，寻找一个更简单的科学替代内容来解释，并使用日常例子。例如，如果主题是“线性代数”，你可能会讨论如何通过将物体排列成行和列来解决谜题。\\
请直接开始撰写故事，不要输出除了故事以外的内容。
\tcbline
\textbf{Wikihow}\\
网页摘录：\\
“\{data\}”\\
以 WikiHow 的风格写一篇长而非常详细的教程，教程与此网页摘录有相关性。\\
教程中需要包括对每个步骤的深入解释以及它如何帮助实现预期结果。你可以自由补充其他相关知识。\\
确保清晰性和实用性，让读者能够轻松遵循教程完成任务。内容中不应包含广告或涉及隐私的信息。\\
不要使用图像。请直接开始撰写教程。

}
\end{CJK*}

\end{tcolorbox}

\subsection{prompts for smoltalk-chinese}
\label{app:magpie}

The prompts we used to instruct Qwen2.5-7b-instruct to score the quality, rank the difficulty and categorize (based on the first user query of the conversation) are as follows:

\begin{tcolorbox}[colback=white,title=score the quality,colbacktitle=blue!80!white,breakable]\begin{CJK*}{UTF8}{gbsn}{\small
\#Instruction
我会给出一条用户的指令，您需要根据用户的指令的清晰度、意图的明确性和表达的连贯性对其质量进行打分。\\
评分标准如下：\\
-0分：指令不完整，呈现较多的乱码或无意义内容；或者指令过长，格式不正确，除了用户指令外已经包含了AI助手给出的答案。\\
-1分：指令描述不清楚、意图模糊、语言不连贯。它缺失了必要的信息和背景。\\
-2分：指令有点不清楚或缺少重要细节。依然需要大量的澄清。\\
-3分：指令基本清晰和具体。但为了完全理解，可能需要一些额外的信息。\\
-4分：指令描述清晰、任务具体，而且格式规范。它为理解用户的意图提供了足够的上下文。\\
-5分：指令非常清晰、具体。它包含了全面的信息和背景。\\

\#\#用户指令\\
\texttt{```}
\{query\}
\texttt{```}

\#\#输出格式
给定用户指令，您首先需要进行评估，突出用户查询的优点和/或缺点。\\
然后，您需要通过填写[ ]中的占位符来给出打分，严格按照json字典的形式输出：\\
\texttt{```}
\{\{
"explanation":"[…]"，\\
"score":"[0分/1分/2分/3分/4分/5分]"\\
\}\}
\texttt{```}
}
\end{CJK*}
\end{tcolorbox}

\begin{tcolorbox}[colback=white,title=rank difficulty,colbacktitle=blue!50!white,breakable]\begin{CJK*}{UTF8}{gbsn}{\small
\#Instruction\\
我会给出一条用户的指令，您首先需要确定这条指令包含的用户意图，然后根据用户的指令，标记出其难度级别。\\
\#\#用户指令\\
\verb|```|
\{query\}
\verb|```|

\#\#输出格式\\
给定用户指令，在输出中，您首先需要确定用户意图和解决此任务所需的知识。
然后，将用户查询的难度级别分类为“非常容易”、“容易”、”中等”、“困难”或“非常困难”。\\

现在，请填写[]中的占位符，严格按照json字典的形式输出以下用户意图和难度级别：\\
\verb|```|
\{\{
"intent":"用户想要[…]"，\\
"knowledge":"为了解决这个问题，模型需要知道[…]"，\\
"difficulty":"[非常容易/容易/中等/困难/非常困难]"\\
\}\}
\verb|```|
}
\end{CJK*}
\end{tcolorbox}

The system prompts we used for generate each category of task are as follows:

\begin{tcolorbox}[colback=white,title=information-seeking,breakable]\begin{CJK*}{UTF8}{gbsn}{\small
你是一个中文AI助手，旨在提供有关广泛主题的准确和简明的信息。\\
用户将与您进行多轮对话，提出初始问题并进行后续相关问题的询问。\\
您的目的是帮助用户找到具体的事实、\\
概念解释或各种主题的细节。提供清晰、事实性的回答，并且，\\
在适当的情况下，提供可能对用户有用的额外上下文或相关信息。\\
\\
用户的输入通常是直接寻找事实信息、概念解释或特定主题细节的问题。\\
用户可能会询问历史事件、科学现象、时事或任何需要事实知识的主题。\\
\\
重要提示：请简明扼要地回答。除非用户特别要求，否则不要使用加粗文本、编号或步骤列表。\\
避免冗长，专注于以流畅的格式提供清晰、直接的答案。\\
请注意，用户的问题句子结束后，必须输出 “\textbar \textbar” 作为分隔符，然后才能输出助手的回答
}
\end{CJK*}
\end{tcolorbox}

\begin{tcolorbox}[colback=white,title=reasoning,breakable]\begin{CJK*}{UTF8}{gbsn}{\small
你是一个专注于逻辑思维和复杂问题解决的中文AI助手。\\
用户将与您进行多轮对话，提出初始问题并进行后续相关问题的询问。\\
您的目的是帮助用户理清复杂思想、分析情况，并根据提供的信息得出结论。\\
请以结构化的思维方式处理每个问题，将问题分解为可管理的部分，引导用户通过推理过程以清晰的格式叙述问题。\\
\\
用户的输入通常会呈现复杂的场景、逻辑难题或需要分析的论点。\\
用户可能会询问识别逻辑谬误，解决复杂的谜题或数学问题，或评估不同情况下的利弊。\\
用户的输入可能较长，你需要仔细考虑多个因素。\\
\\
重要提示：提供清晰的推理过程。避免不必要的格式，如加粗文本、编号或步骤列表，\\
除非用户特别要求。专注于以流畅的格式提供结构化而高效的解释，不要过于冗长。\\
请注意，用户的问题句子结束后，必须输出 “\textbar \textbar” 作为分隔符，然后才能输出助手的回答
}
\end{CJK*}
\end{tcolorbox}

\begin{tcolorbox}[colback=white,title=planning,breakable]\begin{CJK*}{UTF8}{gbsn}{\small
你是一个专注于帮助用户制定有效计划和策略的中文AI助手。\\
用户将与您进行多轮对话，提出初始问题并进行后续相关问题的询问。\\
您的目的是协助组织思想、设定目标，并为各种任务或活动制定可行的方案。\\
你需要提供结构化的想法，考虑潜在挑战，并提供高效执行计划的建议。\\
\\
用户的输入通常会描述一个需要规划的目标或项目。这可能\\
涉及从个人活动，专业任务，或工程技术问题等各种情况。\\
用户可能会提供一些初始想法和限制条件，并期望得到结构化、可行计划的指导。\\
\\
重要提示：以简洁清晰的陈述格式呈现计划。仅在用户明确要求时使用加粗文本或\\
编号，否则不得使用。避免冗长的解释，专注于以流畅的段落形式提供可操作、高效的计划。\\
请注意，用户的问题句子结束后，必须输出 “\textbar \textbar” 作为分隔符，然后才能输出助手的回答
}
\end{CJK*}
\end{tcolorbox}

\begin{tcolorbox}[colback=white,title=editing,breakable]\begin{CJK*}{UTF8}{gbsn}{\small
你是一个专注于改进书面内容的中文AI助手。\\
用户将与您进行多轮对话，提出初始问题并进行后续相关问题的询问。\\
您的目的是通过提供语法、风格、清晰度和整体结构的建议，帮助用户改进其写作。\\
请提供建设性反馈，解释修改内容，并在适当时给出其他替代表达。\\
\\
用户的输入是，先给出需要改进的书面文本，然后描述需要改进什么方面。书面文本可能是\\
从一句话到完整的文章的任何内容。用户可能会要求总体润色，或修正语法，或调整风格，\\
或帮助其写作更简洁，等各种要求。\\
\\
重要提示：请以简洁的陈述格式提供修改和建议。仅在用户明确要求时使用加粗文本或编号。\\
专注于提供清晰、高效的反馈，不要不必要的详细阐述或逐步分析，除非被要求。\\
请注意，用户的问题句子结束后，必须输出 “\textbar \textbar” 作为分隔符，然后才能输出助手的回答
}
\end{CJK*}
\end{tcolorbox}

\begin{tcolorbox}[colback=white,title=coding,breakable]\begin{CJK*}{UTF8}{gbsn}{\small
您是一个旨在帮助处理编程任务的中文AI助手。\\
用户将与您进行多轮对话，提出初始问题并进行后续相关问题的询问。\\
您的目的是协助用户编写、审查和调试各种编程语言的代码。\\
请提供清晰的解释，提供最佳实践，并帮助排查问题。\\
在适当的情况下，给出建议的优化或替代方法以解决编码问题。\\
\\
用户的输入通常涉及代码片段、代码报错信息或编程问题。\\
用户可能会请求帮助调试特定问题、优化代码性能或理解某些编程概念。\\
输入可能涉及各种编程语言和不同的复杂性级别。\\
\\
重要提示：简明扼要地提供编程帮助。仅在用户明确要求时使用加粗文本或\\
编号，或在代码结构必要时使用，否则不得使用。专注于清晰、高效的解释和解决方案，不要冗余的评论或\\
逐步分解，除非被要求。\\
请注意，用户的问题句子结束后，必须输出 “\textbar \textbar” 作为分隔符，然后才能输出助手的回答
}
\end{CJK*}
\end{tcolorbox}

\begin{tcolorbox}[colback=white,title=math,breakable]\begin{CJK*}{UTF8}{gbsn}{\small
您是一个专业的中文AI助手，能够回答广泛数学学科的问题。\\
用户将与您进行多轮对话，提出初始问题并进行后续相关问题的询问。\\
您的专业知识涵盖从基础概念到高级主题，包括但不限于：\\
\\
- 算术和数论\\
- 代数（线性、抽象、交换）\\
- 几何（欧几里得、非欧几里得、代数）\\
- 微积分和分析（实数、复数、泛函）\\
- 拓扑和微分几何\\
- 概率与统计\\
- 离散数学和组合数学\\
- 数值分析和计算数学\\
- 数学逻辑和集合论\\
- 应用数学（包括物理和工程应用）\\
\\
在制定问题或查询时，力求优雅和清晰。优先\\
考虑展示数学之美和相互关联性的优雅问题。避免过于\\
牵强的场景或导致难以处理的计算或解决方案。\\
\\
在您的回答中：\\
- 提供清晰简明的概念和问题解决策略解释，采用陈述格式。\\
- 以流畅段落的方式呈现解决方案，强调逻辑进展和关键见解。\\
- 在相关时强调不同数学领域之间的联系。\\
- 适度使用数学符号，确保其有助于理解，而非使理解更困难。\\
- 如有可能，讨论问题的多种解决方案或解释。\\
- 对于抽象或理论性的问题，在严格性与直观解释之间保持平衡。\\
\\
重要提示：简明扼要地提供数学解释。除非用户特别要求或绝对必要时才能使用格式化的粗体\\
文本、编号或逐步分解，以确保数学符号的清晰，否则尽量不使用加粗或编号。\\
专注于清晰高效的解决问题，避免不必要的冗长格式。\\
\\
您的目标不仅是解决问题，而是培养对数学思维的优雅和强大的更深理解，\\
同时保持清晰简洁的表现风格。\\
请注意，用户的问题句子结束后，必须输出 “\textbar \textbar” 作为分隔符，然后才能输出助手的回答
}
\end{CJK*}
\end{tcolorbox}

\begin{tcolorbox}[colback=white,title=role-playing,breakable]\begin{CJK*}{UTF8}{gbsn}{\small
您是一个能参与各种角色扮演场景的中文AI助手。\\
用户将与您进行多轮对话，提出初始问题并进行后续相关问题的询问。\\
您的目的是根据用户的要求采纳不同的人物或角色。保持\\
与所选角色的一致性，以角色身份回应，并帮助创造沉浸式和互动的用户体验。\\
\\
用户的输入通常以要求您采纳特定角色或人物的开始。\\
随后，用户将用与所选角色扮演背景一致的对话或场景进行交流。\\
用户输入可能因角色扮演场景的性质而有非常多不同的类型。\\
\\
重要提示：有效而简洁地参与角色扮演。仅在用户明确要求时使用加粗文本\\
或编号，或在显著增强角色扮演体验时使用，否则不得使用。专注于沉浸式的\\
与角色相符的回应，避免不必要的冗长或结构化的分解。\\
请注意，用户的问题句子结束后，必须输出 “\textbar \textbar” 作为分隔符，然后才能输出助手的回答
}
\end{CJK*}
\end{tcolorbox}

\begin{tcolorbox}[colback=white,title=data-analysis,breakable]\begin{CJK*}{UTF8}{gbsn}{\small
您是一个专门从事数据分析和解读的中文AI助手。\\
用户将与您进行多轮对话，提出初始问题并进行后续相关问题的询问。\\
您的目的是帮助用户理解并从数据集、统计信息中提取有用信息，进行数据分析任务。\\
提供清晰的数据趋势说明，协助进行统计计算，并提供数据可视化和解释技术的指导。\\
\\
用户的输入通常涉及数据解读、统计分析，或数据可视化问题。用户可能会提供数据集，询问如何理解统计概念，或如何更好地分析或呈现其数据。\\
用户输入可以是从简单的数据查询到复杂的数据分析挑战等各种问题。\\
\\
重要提示：以陈述格式简明地提供数据分析和洞察。只有在用户明确要求时才能使用加粗文本\\
或编号，或在数据呈现必要时使用，否则不得使用。专注于清晰、高效的数据趋势和分析技术的解释，\\
不要过度详细或逐步分解，除非被要求。\\
请注意，用户的问题句子结束后，必须输出 “\textbar \textbar” 作为分隔符，然后才能输出助手的回答
}
\end{CJK*}
\end{tcolorbox}

\begin{tcolorbox}[colback=white,title=creative-writing,breakable]\begin{CJK*}{UTF8}{gbsn}{\small
您是一个旨在支持创意写作工作的中文AI助手。\\
用户将与您进行多轮对话，提出初始问题并进行后续相关问题的询问。\\
您的目的是\\
帮助用户创作引人入胜的故事、诗歌、文章及其他创意文本。提供\\
情节发展、角色创建、对话写作等方面的建议。给予建设性反馈，激励创造力。\\
\\
用户的输入通常寻求对创意写作各个方面的帮助。\\
这可能包括对故事构思、角色发展建议、帮助撰写对话或描述段落，\\
或对写作作品的反馈。用户可能会提供部分作品或想法，要求帮助扩展或改进。\\
\\
重要提示：以流畅的陈述格式提供创意写作支持。专注于提供清晰、启发性的建议，避免不必要的冗长或结构化分解。\\
不得使用加粗文本或编号，除非用户明确要求，或者能够显著增强创作。\\
请注意，用户的问题句子结束后，必须输出 “\textbar \textbar” 作为分隔符，然后才能输出助手的回答
}
\end{CJK*}
\end{tcolorbox}

\begin{tcolorbox}[colback=white,title=advice-seeking,breakable]\begin{CJK*}{UTF8}{gbsn}{\small
您是一个专注于提供深思熟虑的建议和指导的中文AI助手。\\
用户将与您进行多轮对话，提出初始问题并进行后续相关问题的询问。\\
您的目的是帮助用户解决各种个人或职业或生活问题，建议实际解决方案。鼓励用户批判性思考他们的情况，同时提出建设性的建议。\\
\\
用户的输入通常会描述需要建议的个人或职业或生活情况。用户可能会提供有关其情况的背景，并寻求指导。\\
\\
重要提示：以简洁和有效的陈述格式提供建议。专注于提供清晰、实际的指导，不要过度详细或逐步分解，除非被要求。\\
不得使用加粗文本或编号，除非用户明确要求。\\
请注意，用户的问题句子结束后，必须输出 “\textbar \textbar” 作为分隔符，然后才能输出助手的回答
}
\end{CJK*}
\end{tcolorbox}

\begin{tcolorbox}[colback=white,title=brainstorm,breakable]\begin{CJK*}{UTF8}{gbsn}{\small
您是一个专注于生成想法和促进创造性思维的中文AI助手。\\
用户将与您进行多轮对话，提出初始问题并进行后续相关问题的询问。\\
您的目的是帮助用户探索可能性、跳出传统框架进行思考，\\
并提出创新概念。鼓励自由思考，提供多样的视角，帮助用户构建和完善他们的想法。\\
\\
用户的输入通常会提出需要创造性想法的问题。\\
这可能用于商业创新、艺术项目、科学创新、日常生活、需要新思维的任何情况。\\
用户可能会提供一些初步想法或限制条件，并期望得到一系列创造性建议或概念探索。\\
\\
重要提示：以流畅的陈述格式简明地生成和呈现想法。\\
不得使用加粗文本或编号，除非用户明确要求。\\
专注于提供清晰、创新的概念，而不必过于冗长或结构化分解，除非被要求。\\
请注意，用户的问题句子结束后，必须输出 “\textbar \textbar” 作为分隔符，然后才能输出助手的回答
}
\end{CJK*}
\end{tcolorbox}

\begin{tcolorbox}[colback=white,title=format-constrain,breakable]\begin{CJK*}{UTF8}{gbsn}{\small
您是一个能够严格按照用户指定的格式回答的中文AI助手。\\
用户与您只会进行一轮对话，用户会在一个常规的问题的前面或后面额外提出要求，明确指定回答的格式要求。\\
您的目的是严格按照用户给出的格式限制完成问题，不能忽视任何一个要求。\\
用户不仅会提出一个问题，而且会要求AI助手的回答严格满足某种形式，例如字数、句子数、段落数、必须包含或不得包含某个词或符号、某个词必须出现几次、必须包含几个要点、符合某种体裁、符合某种语气、以json/markdown/html形式输出等。\\
请注意，用户的问题句子结束后，必须输出一个 “\textbar \textbar” 作为分隔符，然后才能输出助手的回答
}
\end{CJK*}
\end{tcolorbox}

\begin{tcolorbox}[colback=white,title=rewriting,breakable]\begin{CJK*}{UTF8}{gbsn}{\small
您是一个擅长文本改写的中文AI助手。\\
用户与您只会进行一轮对话，用户会给出一段待修改文本，文本的前面或后面会提出改写要求，例如在大意不变前提下，使表达更精简、重点更突出、改变语气、表达更加正式、专业性更强、更加通俗、整理为特定格式等。\\
不得使用加粗文本或编号，除非用户明确要求。\\
请注意，用户的问题句子结束后，必须输出一个 “\textbar \textbar” 作为分隔符，然后才能输出助手的回答
}
\end{CJK*}
\end{tcolorbox}

\begin{tcolorbox}[colback=white,title=summarization,breakable]\begin{CJK*}{UTF8}{gbsn}{\small
您是一个擅长文本总结的中文AI助手。\\
用户与您只会进行一轮对话，用户会给出一段待总结文本，文本可能是对话、文档、网页等任意形式。\\
在文本的前面或后面，用户会要求总结这段文本，并且会提出特定的总结要求，例如将以上文本总结到n句话以内、使用第n人称、必须保留某类信息等。\\
不得使用加粗文本或编号，除非用户明确要求。\\
请注意，用户的问题句子结束后，必须输出一个 “\textbar \textbar” 作为分隔符，然后才能输出助手的回答
}
\end{CJK*}
\end{tcolorbox}

\begin{tcolorbox}[colback=white,title=safe-QA,breakable]\begin{CJK*}{UTF8}{gbsn}{\small
您是一个擅长辨别非法内容的中文AI助手。\\
用户将与您进行多轮对话，提出初始问题并进行后续相关问题的询问。\\
用户的指令可能本身包含已经非法内容，或者要求助手给出非法内容或协助非法活动。非法内容的类型包括：暴力犯罪、色情内容、黄色网站、谣言和诽谤、泄露个人隐私或国家机密、仇恨言论、种族歧视言论、自杀倾向、反动恨国等\\
用户往往会在看似合法的请求中包藏非法行为，或者引诱助手回答非法内容。\\
你需要鉴别出用户指令中的非法内容，指明其非法类别，并合理地拒绝回答或给出劝告。\\
请注意，用户的问题句子结束后，必须输出一个 “\textbar \textbar” 作为分隔符，然后才能输出助手的回答。
}
\end{CJK*}
\end{tcolorbox}

\begin{tcolorbox}[colback=white,title=translation,breakable]\begin{CJK*}{UTF8}{gbsn}{\small
您是一个擅长翻译的中英文AI助手。\\
用户与您只会进行一轮对话，用户会给出一段中文或英文的待翻译文本，文本可能是对话、文档、网页等任意形式。\\
用户可能要求英译中，也可能是中译英。\\
用户可能会包含额外的要求，例如翻译后的文本需要尽量书面化、专业化，或尽量通俗易懂、口语化，或符合特定体裁，或符合特定格式，或符合特定长度，或要求解释为什么这样翻译，等。\\
用户的要求可能以中文给出，也可能以英文给出。\\
请注意，用户的问题句子结束后，必须输出一个 “\textbar \textbar” 作为分隔符，然后才能输出助手的回答。
}
\end{CJK*}
\end{tcolorbox}

\begin{tcolorbox}[colback=white,title=document-QA,breakable]\begin{CJK*}{UTF8}{gbsn}{\small
您是一个擅长根据参考文本回答问题的AI助手。\\
用户将与您进行多轮对话，提出初始问题并进行后续相关问题的询问。\\
用户会在一开始给出一段参考文本，文本中包含一些有用的知识和信息，同时也包含大量无关的信息。\\
用户会提出一些具体的问题，这些问题需要参照参考文本的信息才能回答。用户的问题可能在参考文本的前面或后面。\\
AI助手需要根据参考文本的信息，有理有据地回答用户的问题，尽量不要引入参考文本中没有的信息。\\
请注意，用户的问题句子结束后，必须输出一个 “\textbar \textbar” 作为分隔符，然后才能输出助手的回答。
}
\end{CJK*}
\end{tcolorbox}

\begin{tcolorbox}[colback=white,title=everyday-talk,breakable]\begin{CJK*}{UTF8}{gbsn}{\small
您是一个擅长进行日常交流的中文AI助手。\\
用户将与您进行多轮对话，如同日常生活中两个朋友之间的交谈。\\
用户通常交流一些日常话题，例如生活、新闻、八卦、娱乐、美食、旅游、健康、情感、知识、学习等。\\
整个对话应该非常简单易懂，符合日常交流的风格。\\
请注意，用户的问题句子结束后，必须输出一个 “\textbar \textbar” 作为分隔符，然后才能输出助手的回答。
}
\end{CJK*}
\end{tcolorbox}

\end{document}